\def\year{2019}\relax
\documentclass[letterpaper]{article} %DO NOT CHANGE THIS
\usepackage{aaai19}  %Required
\usepackage{times}  %Required
\usepackage{helvet}  %Required
\usepackage{courier}  %Required
\usepackage{url}  %Required
\usepackage{graphicx}  %Required
\usepackage{color}
\usepackage{times}
\usepackage{url}
\usepackage{latexsym}
\usepackage{amssymb}
\usepackage{graphicx}
\usepackage{epsfig}
\usepackage{algorithm}
\usepackage{algorithmic}
\usepackage{amsmath}
\usepackage{caption}
\usepackage{subcaption}
\usepackage{epstopdf}
\usepackage{fancybox}
\usepackage{blindtext}

\usepackage{tikz}
\usetikzlibrary{shapes, arrows, positioning, decorations.markings}

\definecolor {processblue}{cmyk}{0.96,0,0,0}
\tikzstyle{int}=[draw, fill=blue!20, minimum size=2em]
\tikzstyle{init} = [pin edge={to-,thin,black}]

\usetikzlibrary{shapes}
\usetikzlibrary{fit}
\usetikzlibrary{chains}
\usetikzlibrary{arrows}

\tikzstyle{plate} = [draw, rectangle, rounded corners, fit=#1]
\tikzstyle{wrap} = [inner sep=0pt, fit=#1]

\tikzstyle{caption} = [node distance=0] %
\tikzstyle{bottom plate caption} = [caption, node distance=0, inner sep=0pt,
below left=-5pt and 0pt of #1.south east] %

\tikzstyle{top plate caption} = [caption, node distance=0, inner sep=0pt,
below left=0pt and 0pt of #1.north east] %

\newtheorem{hypo}{Hypothesis}
\newtheorem{lem}{Lemma}

\begin{document}
% The file aaai.sty is the style file for AAAI Press 
% proceedings, working notes, and technical reports.
%

\title{MixUp as Locally Linear Out-Of-Manifold Regularization}
%\title{Locally Linear Out-Of-Manifold Regularization}
%\author{}
\author{
  Hongyu Guo \\
  National Research Council Canada\\
  1200 Montreal Road, Ottawa \\
  \texttt{hongyu.guo@nrc-cnrc.gc.ca} \\
  %% examples of more authors
   \And
   Yongyi Mao \\
   School of Electrical Engineering \& Computer Science\\
University of Ottawa, Ottawa, Ontario\\
   %Address \\
   \texttt{yymao@eecs.uottawa.ca} \\
   \AND
   Richong Zhang \\
   BDBC, School of Computer Science and Engineering\\
   Beihang University, Beijing, China \\
%   Address \\
   \texttt{zhangrc@act.buaa.edu.cn} 
}

\maketitle

\begin{abstract}
MixUp~\cite{mixup17} is a recently proposed data-augmentation scheme, which linearly interpolates a random pair of training examples and correspondingly the one-hot representations of their labels. Training deep neural
networks with such additional data is shown capable of significantly improving the predictive accuracy of the current art. The power of MixUp, however, is primarily established empirically and its working  and  effectiveness have not been explained in any depth.  In this paper, we  develop an understanding for MixUp as a form of ``out-of-manifold  regularization'', which imposes certain ``local linearity'' constraints on the model's input space beyond the data manifold.  This analysis enables us to identify a limitation of MixUp, which we call ``manifold intrusion''.  In a nutshell, manifold intrusion in MixUp is a form of under-fitting resulting from conflicts between the 
synthetic labels of the mixed-up examples and the labels of original training data. Such a phenomenon usually happens when the parameters controlling the generation of mixing policies are not sufficiently fine-tuned  on the training data.  To address this issue, we propose a novel adaptive version of MixUp, where the mixing policies are automatically learned from the data using an additional network and objective function designed to avoid manifold intrusion. The proposed regularizer, AdaMixUp, is empirically evaluated on several 
benchmark datasets. Extensive experiments demonstrate that AdaMixUp improves upon MixUp when applied to the current art of deep classification models.

\end{abstract}

\section{Introduction}
\label{intr}
Deep learning techniques have  achieved profound success in many challenging real-world applications, including image recognition~\cite{Krizhevsky:2012:ICD:2999134.2999257,DBLP:journals/corr/HeZR016}, speech recognition~\cite{38131,DBLP:journals/corr/abs-1303-5778}, and machine translation~\cite{DBLP:journals/corr/SutskeverVL14,DBLP:journals/corr/BahdanauCB14}, amongst many others~\cite{DBLP:conf/nips/GoodfellowPMXWOCB14,DBLP:journals/nature/SilverHMGSDSAPL16,DBLP:journals/corr/KipfW16}. 
A recently proposed such technique, MixUp~\cite{mixup17}, is a simple and yet very effective  data-augmentation approach to enhance the performance of deep classification models.  Through linearly interpolating random data sample pairs and their training targets, 
MixUp generates a synthetic set of examples and use  these examples to augment the training set.
In ~\cite{mixup17}, MixUp is shown to dramatically improve the predictive accuracy of the current art of deep neural networks. 

Despite its demonstrated effectiveness, the power of MixUp is mostly established empirically. To date, the working of MixUp has not been well explained. In addition, the mixing (i.e., interpolation) policies in MixUp are controlled by a global hyper-parameter $\alpha$, which needs to be tuned by trial and error on the data set. This on one hand makes MixUp inconvenient to use, and on the other hand, adds to the mystery what role such a parameter controls and how to tune it properly.

The study of this current paper is motivated by developing a deeper understanding of MixUp, specifically pertaining to why it works. To that end, we formulate MixUp as a new form of regularization. Specifically,  we categorize regularization schemes as {\em data-independent} regularization and {\em data-dependent} regularization. Data-independent regularization imposes constraints on the model without exploiting the structure (e.g., the distribution) of data; typical examples include penalizing various forms of the norm of the network parameters (e.g., weight decay~\cite{DBLP:conf/nips/HansonP88}) or dropout~\cite{Srivastava:2014:DSW:2627435.2670313}. Data-dependent regularization constrains the parameter space of the model in a way that depends on the structure of the data; previous examples of such schemes include, data augmentation schemes~\cite{Simard:1998:TIP:645754.668381,726791,SimonyanZ14a}, adversarial training schemes (e.g., \cite{Goodfellow-advExamples}), and some information bottleneck based regularization schemes (e.g., \cite{DVIB}). In this paper, we show that MixUp is also a data-dependent regularization scheme in the sense that the imposed constraints on the model explicitly exploit the data distribution. But different from all previous data-dependent regularization schemes, MixUp imposes its constraints by making use of the regions in the input space of the model that are outside of the data manifold. Identifying the constraints imposed by MixUp with a set of ``locally linear'' constraints, we call such a data-dependent regularization scheme a ``locally linear out-of-manifold'' regularization scheme.

%Unlike the conventional regularization schemes (such as weight decay~\cite{DBLP:conf/nips/HansonP88}, dropout~\cite{Srivastava:2014:DSW:2627435.2670313}, %stochastic depth~\cite{DBLP:journals/corr/HuangSLSW16}, 
%and batch normalization~\cite{DBLP:journals/corr/IoffeS15}), which are data-independent, MixUp makes explicit use of the training data. Also, different from data augmentation schemes~\cite{Simard:1998:TIP:645754.668381,726791,SimonyanZ14a,DBLP:journals/corr/abs-1807-10251}, which may be argued also as a regularization scheme, MixUp constrains the model on the input space beyond the data manifold. 

\begin{figure}[h]
\caption{Left: Linearly interpolated images (top row) from the original images (bottom two rows). Right: Performance of MixUp  
on a reduced  Cifar100 data set (reduced to containing 20\% of data samples)
vs various values  of $\alpha$.  
}
	\centering
\includegraphics[width=3.3in]{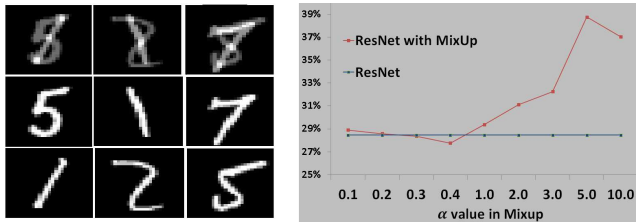}   
	\label{fig:cifar100mess}
\end{figure}

Under this perspective, we identify an intrinsic problem in MixUp, which we  refer to as ``manifold intrusion''.  Briefly, manifold intrusion occurs when a mixed example collides with a real example in the data manifold, but is given a soft label that is different from the label of the real example. Figure~\ref{fig:cifar100mess} (left) shows some examples of manifold intrusion. In the figure, the mixed images (top row) look close to a hand-written  ``8'' (which indeed live in the data set) and yet MixUp assigns a soft-label that is not ``8''. For example, the soft-label of the top-left image is given as the half-half mix of labels ``1'' and ``5''. 
We explain in this work that manifold intrusion necessarily results in under-fitting and degradation of the model performance. This is confirmed by our experiments, where we see MixUp with an inappropriate hyper-parameter $\alpha$ actually deteriorates the model performance (Figure~\ref{fig:cifar100mess}, right). Indeed, with MixUp, one has to select hyper-parameter $\alpha$ carefully via trial and error, in order to enhance the base model.

Built on our understanding and motivated by the 
potential manifold intrusion in MixUp, we then propose a generalized and adaptive version of MixUp, which we call AdaMixUp. On one hand, AdaMixUp extends the mixing policies in the standard MixUp from 2-fold to higher-fold; more importantly  it adaptively learns good policy regions from the data and automatically avoids manifold intrusion. Our experiments on several  image classification tasks show that AdaMixUp significantly improves the current art of deep classification models and outperforms the standard MixUp by a large margin.

\section{Preliminaries}
In a standard classification setting, let ${\cal X}$ denote the vector space in which each example $x$ lives.  Throughout the paper, the elements of ${\cal X}$ will be treated as vectors and all vectors in this paper are taken as column vectors. 

Let ${\cal Y}$ denote the set of all class labels. The objective of our classification problem is to develop a classifier which assigns every example a label in ${\cal Y}$.

It is worth noting however that not every $x\in {\cal X}$ is a valid example. That is, some $x$ cannot be associated to  a label in ${\cal Y}$. Specifically, the set of all valid examples is only a subset of ${\cal X}$, which we refer to as the {\em data manifold}, or in short, the manifold, and denote it by ${\cal M}$. 
%(Let the distribution $p_X$ on ${\cal X}$ be the distribution of valid examples. Then ${\cal M}$ is essentially the support of $p_X$.) 

The following hypothesis is adopted for nearly every classification problem.
\begin{hypo}[Basic Hypothesis]
\label{ax:g}
Each $x\in {\cal M}$ has a {\em unique} label in ${\cal Y}$. We use $g(x)$ to denote the label of $x$, where $g$ is a function mapping ${\cal M}$ onto ${\cal Y}$. We stress that on ${\cal X}$ outside ${\cal M}$, $g$ is not defined.
\end{hypo}

In the classification problem, we are given a subset ${\cal D}\subset {\cal M}$ as the training examples, where for each $x\in {\cal D}$, $g(x)$ is given.

A distribution over a set of $m$ elements will be denoted by a vector in ${\mathbb R}^m$. It is well known that the set of all such distributions form the {\em probability simplex} in ${\mathbb R}^m$, which we will denote by ${\mathbb S}_m$. The set of all distributions on ${\cal Y}$ is obviously a probability simplex, but we will give it a separate notation ${\cal P}({\cal Y})$ for distinction. Specifically, each distribution in ${\cal P}({\cal Y})$ is regarded as a ``soft label'', whereas each distribution in other probability simplexes potentially serves as a set of coefficients, which will be used to mix training examples via a convex combination. For example, a Bernoulli distribution $[\alpha, 1-\alpha]^T$ as a distribution in ${\mathbb S}_2$ may potentially be used to mix two examples $x, x'\in {\cal X}$ by $\alpha x+ (1-\alpha) x'$. Thus the distributions in these simplexes will be referred to as a {\em mixing policy}, or simply {\em policy}. Any subset $\Lambda$ of these probability simplex with non-zero Lebesgue measure will then be called a {\em policy region}. A distribution or policy will be called {\em degenerate} if it is a distribution containing a single non-zero probability mass.

Let ${\cal F}({\cal X}, {\cal Y})$ denote the space of all functions mapping ${\cal X}$ to ${\cal P}({\cal Y})$. A {\em discriminative classification model} (such as a neural network model) that outputs a predictive distribution over ${\cal Y}$ for an example $x\in {\cal X}$ can be characterized as a subset ${\cal H} \subset {\cal F}({\cal X}, {\cal Y})$。

Given the model ${\cal H}$, the objective of learning is then finding a member function $H\in {\cal H}$ which hopefully, for each $x\in {\cal M}$, returns a predictive distribution that puts highest probability on 
$g(x)$.  Usually such an $H$ is found by optimizing a well-defined loss
function ${\cal L}_{\cal D}(H)$ over all $H$ in the space ${\cal H}$ 
\begin{equation}
\label{eq:basicOpt}
\widehat{H}:=\arg\min_{H\in {\cal H}} {\cal L}_{\cal D}(H).
\end{equation}
We note that we subscript the loss with ${\cal D}$ to emphasize its dependency on the training data ${\cal D}$.

\section{Regularization}

It is well known that when the space ${\cal H}$ is large (for example, high dimensional), or when the model is too complex, the model tends to overfit and generalize poorly.
The main techniques to cure such a problem is {\em regularization}. 

Regularization, in our view, may refer to any technique that effectively reduces the model capacity or reduces the space ${\cal H}$. In this paper, we wish to divide regularization into two categories: {\em data-dependent} and {\em data-independent}.

{\em Data-independent regularization} is the usual notion of regularization. Typical such techniques directly impose certain constraint, say, $C(H)<A$, on the functions $H\in {\cal H}$.  This turns the optimization problem (\ref{eq:basicOpt})  to a constrained optimization problem, 
which, under the Lagrangian formulation, can be expressed by
\begin{equation}
\label{eq:regOpt2}
\widehat{H}:=\arg\min_{H\in {\cal H}} \left\{ {\cal L}_{\cal D}(H) + \beta C(H) \right\}
\end{equation}

For example, the weight-decay regularization is a data-independent regularization, which essentially imposes a L2-norm constraint on the model parameters.
Dropout~\cite{Srivastava:2014:DSW:2627435.2670313} can be regarded as another example of such a regularizer, which can be understood as reducing the model capacity in a Bayesian style~\cite{DropoutAsBayesian} and which has an equivalent loss function similar to the form of (\ref{eq:regOpt2}).  A key feature in data-independent regularization is that the training data do not enter the regularization term $C(H)$ in such a scheme.

{\em Data-dependent regularization} imposes constraints or makes additional assumptions on ${\cal H}$ with respect to the training data ${\cal D}$. 

The most well known such techniques are data augmentation.
Specifically, in a data augmentation scheme, another set ${\cal D}' \subset {\cal X}$, {\em believed to be inside ${\cal M}$} but beyond the training set ${\cal D}$, are introduced for training. For example, if ${\cal D}$ is a set of images, ${\cal D}'$ can be obtained by rotating each image in ${\cal D}$; the label $g(x)$ of an image $x$ in ${\cal D}'$ is taken as the label of the image $x$ before rotation. Data augmentation turns the optimization problem (\ref{eq:basicOpt}) into

\begin{equation}
\label{eq:dataAugment}
\widehat{H}:=\arg\min_{H\in {\cal H}} \left\{ {\cal L}_{\cal D}(H) + {\cal L}_{{\cal D}'}(H) \right\}
\end{equation}
where ${\cal L}_{{\cal D}'}$ is the same loss function but on data ${\cal D}'$. Aligning  (\ref{eq:dataAugment}) with 
(\ref{eq:regOpt2}), one can equivalently regard data augmentation as imposing a constraint on the space ${\cal H}$ (hence a regularization scheme), and such a constraint, depending on the data set ${\cal D}$, is precisely due to the hypothesis that ${\cal D}' \subset {\cal M}$ and $g(\cdot)$ is rotation-invariant.

Another data-dependent regularization scheme is adversarial training (see, e.g., \cite{Goodfellow-advExamples}). In such a scheme, for each training example $x$, an adversarial example $x'$ is found, and it is assumed that $x'$ has the same label as $x$.  The adversarial examples are then also used for training purpose. Although delicate procedures have been proposed to find adversarial examples, the found adversarial examples are used in the same way as the additional data set ${\cal D}'$ in data augmentation.

In both data augmentation and adversarial training, the 
constraints imposed on the label function $g$ is in the data manifold; the constraints essentially specify how function $g$ should behave on additional points in the manifold ${\cal M}$. Next we will argue that MixUp may be viewed as another data-dependent regularization scheme, which introduces constraints on $g$ outside the manifold ${\cal M}$.

\section{Locally Linear Out-of-Manifold Regularization}

For each $y\in {\cal Y}$, let $\delta_y\in {\cal P}({\cal Y})$ be the single-point-mass (namely, degenerate) distribution on ${\cal Y}$ which puts all the probability mass on $y$. Now we associate with $g$ a function $G: {\cal X}\rightarrow {\cal P}({\cal Y})$ satisfying
\begin{equation}
\label{eq:G}
G(x):=\delta_{g(x)}, {\rm for ~any} ~x\in {\cal M}.
\end{equation}
Obviously, on the data manifold ${\cal M}$, $G$ is simply a representation of $g$ in ${\cal F}({\cal X}, {\cal Y})$.  Additionally any ${\cal G}$ satisfying the above equation is as perfect as $g$ for any valid example.  Thus, any such function $G$ can be taken as a ``ground-truth'' classifier. 

Then the classification problem can be reformulated as 
constructing a model ${\cal H}\subset
{\cal F}({\cal X}, {\cal Y})$ and finding a member $H\in {\cal H}$ that well approximates a function $G$ satisfying (\ref{eq:G}).

Noting that the condition (\ref{eq:G}) on $G$ is only within ${\cal M}$, and it imposes no condition on the space ${\cal X}$ outside ${\cal M}$. However, the functions in ${\cal H}$ are defined on ${\cal X}$, well beyond ${\cal M}$.  Thus, if one chooses to regularize the model in a data-dependent manner, there is a lots of degrees of freedom beyond ${\cal M}$ which one can exploit.  We will show that the MixUp strategy presented in \cite{mixup17} can be seen as such a technique.

To begin, for any subset ${\Omega}\subseteq {\cal X}$, we will use ${\Omega}^{(k)}$ to denote the set of all $k$-column matrices in which each column is a point (vector) in $\Omega$.  This also defines notations 
${\cal M}^{(k)}$ and ${\cal D}^{(k)}$; however with these latter two notations, we require
each matrix in ${\cal M}^{(k)}$ and ${\cal D}^{(k)}$ satisfy an additional condition, namely, {\em the labels of the columns of the matrix must be all distinct}.

Let $k$ be a given integer not less than 2. Let $\Lambda\subseteq {\mathbb S}_k$ be a policy region, and ${\bf X}\in {\cal X}^{(k)}$. We say that a function $F\in {\cal F}({\cal X}, {\cal Y})$ is {\em $\Lambda$-linear with respect to ${\bf X}$} if for any $\lambda \in \Lambda$ 
\[
F({\bf X}\lambda) = F({\bf X})\lambda
\]
where $F(\cdot)$, when having its domain extended to matrices, acts on matrix  ${\bf X}$ column-wise.  We also refer to such a notion of linearity as a {\em $k$-fold local linearity}. We note that if we allow $\Lambda$ to ${\mathbb R}^k$, then ${\mathbb R}^k$-linearity with respect to all  ${\bf X}\in {\cal X}^{(k)}$ for every integer $k\ge 1$ implies the standard linearity.  

\subsection{Standard MixUp}

\begin{hypo}[Standard MixUp Hypothesis] 
\label{hypo:standardMixUp}
There exists a policy region $\Lambda \subset {\mathbb S}^2$ such that for any matrix ${\bf X}\in {\cal M}^{(2)}$, the function ${G}$ is $\Lambda$-linear with respect to ${\bf X}$. 
\end{hypo}

We note that this hypothesis only assumes a fold-2 local linearity and a global choice of $\Lambda$. If Hypothesis \ref{hypo:standardMixUp} holds and  one is given $\Lambda$, then one can implementing the hypothesis as a constraint on ${\cal H}$, by imposing on the optimization problem the following constraint.

\centerline{
\shadowbox
{
\parbox[t]{7.5cm}{
Standard MixUp Constraint:\\
Every $H\in {\cal H}$ is $\Lambda$-linear w.r.t. every ${\bf X}\in {\cal D}^{(2)}$, 
}
}
}

This gives rise a regularization scheme, which can be implemented by repeating the following process: draw a random pair data points\footnote{Although we have required in the definition of ${\cal D}^{(2)}$ that two points in matrix ${\bf X}$ should have different labels, this needs not to be rigorously followed in 
implementation. Relaxing this condition in fact is expected to decrease the chance of manifold intrusion. Similar comments apply to the AdaMixUp later.} in ${\cal D}$ to form ${\bf X}$, and draw a random $\lambda$ from $\Lambda$; take ${\bf X}\lambda$ as an additional training example and $G({\bf X})\lambda$ as its training target.

This scheme, generating more training examples, is essentially the standard MixUp scheme \cite{mixup17}.

In this paper, we however point out that one needs to caution with such a regularization scheme. 

\begin{lem} 
\label{lem:standardMixUpFail}
Let $\Lambda$ be a policy region in ${\mathbb S}_2$ and ${\bf X} \in {\cal M}^{(2)}$.
%and ${\bf X}:=[x_1, x_2] \in {\cal M}^{(2)}$ with $g(x_1)\neq g(x_2)$. 
If there is a non-degenerate $\lambda \in \Lambda$ with ${\bf X}\lambda \in {\cal M}$, then Hypothesis \ref{hypo:standardMixUp} can not hold with this choice $\Lambda$\footnote{More precisely, in order to make the hypothesis fail, instead of requiring a single non-degenerate $\lambda$, we in fact require a subset of 
$\Lambda$ having non-zero measure. But we abbreviate it here for simplicity. Similar comments apply to Lemma. \ref{lem:generalMixUpFail}. \label{footnote:intrudeStdMixUp}
}.
\end{lem}

%This result is easy to see: if the hypothesis held with such an $\Lambda$, Hypothesis \ref{ax:g} would be violated. 

\noindent {\em Proof:} The condition of the lemma, stated more precisely in Footnote \ref{footnote:intrudeStdMixUp} suggests that there is a region $\Lambda' \subseteq \Lambda$ with non-zero Lebesgue measure such that every $\lambda \in \Lambda'$ is non-degenerate and ${\bf X}\lambda \in {\cal M}$. The fact that 
${\bf X}\lambda \in {\cal M}$ suggests, by Hypothesis \ref{ax:g}, that 
$g({\bf X}\lambda)\in {\cal Y}$, or ${\cal G}
({\bf X}\lambda)$ is a single-point-mass distribution for every $\lambda \in \Lambda'$. 

To induce a contradiction, suppose that Hypothesis \ref{hypo:standardMixUp} holds for $\Lambda$, namely, $G$ is $\Lambda$-linear with respect to 
every ${\bf X}\in {\cal M}^{(2)}$. Then
${\cal G}({\bf X}\lambda) = {\cal G}({\bf X}) \lambda$ for every $\lambda \in \Lambda' \subseteq \Lambda$. Since the two columns (points) in ${\bf X}$ have different labels,  
the two columns of $G({\bf X})$ must be two different one-hot vectors (degenerate distributions). But every $\lambda\in \Lambda'$ is  non-degenerate; as a consequence, every $G({\bf X})\lambda$ must be non-degenerate. This contradicts the previous argument that ${\cal G} ({\bf X}\lambda)$ is a single-point-mass. Therefore Hypothesis \ref{hypo:standardMixUp} fails on a region in $\Lambda$ with 
non-zero measure. \hfill $\Box$

Geometrically, the failure of Hypothesis \ref{hypo:standardMixUp} in this lemma is due to that the mixing $x_1$ and $x_2$ by $\lambda$ causes the mixed point ${\bf X}\lambda$ to ``intrude'' into ${\cal M}$ and ``collide'' with a point, say, $x'$ in ${\cal M}$. 
We call such a phenomenon {\em manifold intrusion}.
Note that in this case, ${\bf X}\lambda$ and $x'$ are  in fact the same point, but they have different labels. Obviously, when such intrusion occurs, regularization using ${\bf X}$ and $\lambda$ will contradict with the original training data. This essentially induces a form of {\em under-fitting}. 

\begin{figure}[h]
\caption{
\label{fig:venn}
Effect of MixUp on constraining ${\cal H}$}
	\centering
\includegraphics[width=3.31in]{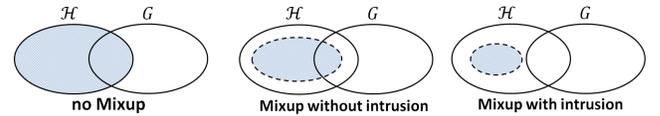}         
\end{figure}

Figure \ref{fig:venn} depicts the effect of MixUp on constraining the space ${\cal H}$, where the region ${\cal G}$ denotes the space of functions in ${\cal F}({\cal X}, {\cal Y})$ satisfying (\ref{eq:G}). 

When the space ${\cal H}$ is large, there is a large region of ${\cal H}$ satisfying (\ref{eq:G}). This gives a large intersection of ${\cal H}$ and ${\cal G}$, each member of which is a perfect classifier on the training set but performing questionably on the testing set. Thus the model without MixUp tends to result in large prediction variances or over-fitting (left figure). 

When MixUp is applied and manifold intrusion does not occur (middle figure), the MixUp constraint effectively reduces the space ${\cal H}$. Since the MixUp constraint does not conflict (\ref{eq:G}) (at least explicitly), the reduced ${\cal H}$ would contain members compatible with (\ref{eq:G}). Thus the intersection of ${\cal G}$ and ${\cal H}$ reduces while remaining non-empty. Each member in the intersection is a perfect classifier on the training set and the mixed examples.
But the significantly reduced intersection gives rise to significantly lower prediction variances on the testing set, thereby serving to regularize the model and reduce over-fitting. 

When the application of MixUp results in manifold intrusion (right figure), there is no member in ${\cal H}$ that satisfies both the MixUp constraint and the condition (\ref{eq:G}). The intersection of ${\cal H}$ and ${\cal G}$ then becomes empty. In this case, no member in ${\cal H}$ can fit both the original training data and the (intruding) mixed examples. This gives rise to under-fitting and high prediction biases.

From the view point of restricting the space ${\cal H}$, one wishes to make $\Lambda$ as large as possible.  But this lemma sets a limit on the possible $\Lambda$ that makes Hypothesis \ref{hypo:standardMixUp} hold. Due to this lemma, we need to assure that the $\Lambda$ we choose does not cause such an ``intrusion''.  

Let $\Lambda^*$ denote the largest policy region in ${\mathbb S}_2$ such that manifold intrusion does not occur.  In the original MixUp, recall that inappropriate hyper-parameter result in no gain and only negative impact. We believe that much of this is due to manifold intrusion. On the other hand, the search of good hyper-parameter $\alpha$ in the original MixUp can be seen as searching for $\Lambda^*$ by trial and error.

In this paper, we generalize the standard MixUp to regularization with higher-fold local linearities, where $\Lambda^*$ (more precisely its generalization) is learned from the data.  

\subsection{Adaptive MixUp (AdaMixUp)}
%\subsection{Generalized MixUp}

\begin{hypo}[Generalized MixUp Hypothesis] 
\label{hypo:generalMixUp}
For any integer $k$ not less than $2$ and not greater than some $k_{\rm max}$,  and for any matrix ${\bf X}\in {\cal M}^{(k)}$, there exists a policy region $\Lambda \subset {\mathbb S}^k$ such that the function $G$ is $\Lambda$-linear w.r.t. ${\bf X}$. Note that here $\Lambda$ depends on ${\bf X}$, which we will denote by $\Lambda({\bf X})$.
\end{hypo}

If this hypothesis holds, and if for every $k\in \{2, 3, \ldots, k_{\rm max}\}$ and every ${\bf X}\in {\cal M}^{(k)}$, $\Lambda({\bf X})$ in the hypothesis is given, then the hypothesis can be implemented as the following constraint on ${\cal H}$.

\centerline{
\shadowbox
{
\parbox[t]{7.5cm}{
Generalized MixUp Constraint:\\
For every $k\in \{2, 3, \ldots, k_{\rm max}\}$, every ${\bf X}\in {\cal D}^{(k)}$ and every $\Lambda({\bf X})$, every $H\in {\cal H}$ is $\Lambda({\bf X})$-linear w.r.t. ${\bf X}$. 
}
}
}

Imposing such a constraint in training provides another regularization scheme, which will be one ingredient of the proposed AdaMixUp scheme.  We will present AdaMixUp in the next section, after developing its other ingredients.

\begin{lem}
\label{lem:generalMixUpFail} Let $k\in \{2, 3, \ldots, k_{\rm max}\}$ and 
${\bf X}\in {\cal M}^{(k)}$.
%${\bf X}=[x_1, x_2, \ldots, x_k] \in {\cal M}^{(k)}$ where $g(x_1), g(x_2), \ldots, g(x_k)$ are not all the same. 
Suppose that $\Lambda$ is a policy region in ${\mathbb S}_k$. If there is a non-degenerate $\lambda\in \Lambda$ with ${\bf X}\lambda \in {\cal M}$, then Hypothesis \ref{hypo:generalMixUp} with 
$\Lambda({\bf X}) = \Lambda$ can not hold.
\end{lem}

This lemma parallels Lemma \ref{lem:standardMixUpFail} and can be proved similarly.  Lemma \ref{lem:generalMixUpFail} similarly suggests that when the imposed constraint results in intrusion into the manifold ${\cal M}$, Hypothesis \ref{hypo:generalMixUp} necessarily fails.  The lemma also sets a limit for each $\Lambda({\bf X})$. For each ${\bf X}$, we will denote the largest $\Lambda({\bf X})$ by $\Lambda^*({\bf X})$.  The following result immediately follows.

\begin{lem}
For any ${\bf X}\in {\cal M}^{(2)}$, $\Lambda^* \subseteq \Lambda^*({\bf X})$.
\end{lem}

\noindent {\em Proof:} By the definitions of $\Lambda^*$ and
$\Lambda^*({\bf X})$, $\Lambda^* = \bigcap_{{\bf X}\in {\cal M}^{(2)}} \Lambda^* ({\bf X})$. Thus $\Lambda^* \subseteq 
\Lambda^*({\bf X})$.  \hfill $\Box$

In reality, one expects that $\Lambda^*$ is strictly contained in $\Lambda^*({\bf X})$. That is, even when we take $k_{\rm max}=2$, the Generalized MixUp Hypothesis imposes a stronger constraint than the Standard MixUp Hypothesis, and hence can provide a stronger regularization without causing under-fitting. When $k_{\rm max} > 2$, the Generalized MixUp Hypothesis imposes additional constraints, which further regularizes the model and improves generalization.

\subsubsection{AdaMixUp}

From here on, we stay in the Generalized MixUp regime, namely assuming Hypothesis \ref{hypo:generalMixUp} holds.  We now explore the possibility of learning $\Lambda({\bf X})$ for each ${\bf X}$. 

Suppose that each $\Lambda^* ({\bf X})\subset {\mathbb S}_k$ can be sufficiently parametrized by $\pi_k({\bf X})$ in some space $\Pi$.  Then $\pi_k(\cdot)$ can be regarded as a function mapping ${\cal M}^{(k)}$ to $\Pi$.  Then learning each $\Lambda({\bf X})$ reduces to learning the function $\pi_k(\cdot)$.  We now consider using a neural network to represent the function $\pi_k$. This network will be referred to as a {\em policy region generator}.
%, and its parameter will be denoted by $\theta$.  
With a slight abuse of notation, we will use $\pi_k({\bf X})$ to denote any candidate choice of $\Lambda^*({\bf X}) \subset {\mathbb S}_k$ generated by the network $\pi_k(\cdot)$.

To train the networks $\{\pi_k\}$, consider the construction of another network $\varphi(\cdot)$. The network takes any $x\in {\cal X}$ as input and outputs a predictive distribution on whether the $x$ lies in manifold ${\cal M}$.

This network, which we refer to as an  {\em intrusion discriminator}, aims at distinguishing elements inside the manifold ${\cal M}$ and those outside.  By assuring that the virtual examples obtained by mixing with the policies in $\pi_k({\bf X})$ are outside the manifold and the original examples are inside the manifold, the supervising signal trains both the networks $\{\pi_k\}$ and $\varphi$. More details are given below.

For a given $x\in {\cal X}$, let $p(\cdot|x; \varphi)$ be the predictive distribution on $\{1, 0\}$ generated by network $\varphi$, where $1$ denotes $x$ is outside ${\cal M}$. The loss function for training $\{\pi_k\}$ and $\varphi$ is given by the following cross-entropy loss, which we call the ``intrusion loss'':
\begin{equation}
\begin{split}
{\cal L}_{\rm intr}: = & \frac{1}{k_{\rm max}\!\!\!-1}
\sum\limits_{k=2}^{k_{\rm max}}{\mathbb E}_{{\bf X}\sim {\cal D}^{(k)}, \lambda \sim \pi_k({\bf X})} \log p(1|{\bf X}\lambda; \varphi)\\
& + {\mathbb E}_{x \sim {\cal D}} \log p(0|x; \varphi)
\end{split}
\end{equation}

This intrusion loss ${\cal L}_{\rm intr}$, depending on both $\pi_k$'s and $\varphi$  will be responsible for providing training signals for $\pi_k$'s and $\varphi$.

Given $\pi_k$'s, let ${\cal D}'$ be the set of all the virtual examples ${\bf X}\lambda$ that can be obtained by drawing an ${\bf X}$ from ${\cal D}^{(k)}$ for some $k$ and mixing with some policy $\lambda$ in region $\pi_k({\bf X})$. Let the ``MixUp Loss'' ${\cal L}_{{\cal D}'}$ be the average cross-entropy loss for the virtual examples ${\bf X}\lambda$ in ${\cal D}'$ with respect to their respective soft labels $G({\bf X})\lambda$. Then we arrive at an overall loss function 
\begin{equation}
\label{eq:lossAdaMixUp}
{\cal L}_{\rm total}: = 
{\cal L}_{\cal D}(H) + {\cal L}_{{\cal D}'}(H, \{\pi_k\}) + 
{\cal L}_{\rm intr}(\{\pi_k\}, \varphi)
\end{equation}

Training the networks $H$, $\{\pi_k\}$ and $\varphi$ on ${\cal L}_{\rm total}$ then gives rise to the proposed AdaMixUp framework.   Comparing the loss 
in (\ref{eq:lossAdaMixUp}) with the objective function in (\ref{eq:regOpt2}), one can equate ${\cal L}_{{\cal D}'}(H, \{\pi_k\}) + 
{\cal L}_{\rm intr}(\{\pi_k\}, \varphi)$ in (\ref{eq:lossAdaMixUp}) with 
regularization term $\beta C(H)$ in  (\ref{eq:regOpt2}). Specifically, here ${\cal L}_{{\cal D}'}(H, \{\pi_k\})$ serves to regularize the model and ${\cal L}_{\rm intr}(\{\pi_k\}, \varphi)$ serves to prevent the model from 
``over-regularization'', namely, intrusion or under-fitting.  This justifies AdaMixUp (and the standard MixUp) as a regularization technique. The difference between AdaMixUp  (\ref{eq:lossAdaMixUp}) and the standard regularization 
(\ref{eq:regOpt2}) is however two fold: first, unlike the standard regularization which is data-independent, AdaMixUp depends on the dataset; on the other hand, AdaMixUp contains parameters,  which are adaptively learned from the dataset ${\cal D}$.

Given that each $\pi_k$ has sufficient capacity and properly trained, in theory $\pi_k({\bf X})$ can approximate $\Lambda^*({\bf X})$ well. Then if the ${\varphi}$ also has sufficient capacity, the intrusion loss ${\cal L}_{\rm intr}$ can be driven to near zero. In practice however, one usually need to reduce the capacities of $\pi_k$'s and $\varphi$. Then the intrusion loss may be seen as an approximate measure of the extent to which the data allows for mix up without causing intrusion under the architectural choices of $\pi_k$'s and $\varphi$.

% \begin{lem}
% \label{lem4}
% Suppose that Hypothesis \ref{hypo:generalMixUp} holds. Then there exists a finite-capacity networks $\pi_k$ for every $k=2, 3, \ldots, k_{\rm max}$, and a finite-capacity network $\varphi$ such that ${\cal L}_{\rm intr}$ can be made arbitrarily close to 0 and $\pi_k({\bf X})$ gives rise to $\Lambda^*({\bf X})$.  
% \end{lem}

% \noindent {\em Proof:} 
% %For each class label $y$, let ${\cal M}(y)$ denote the manifold of data having class label $y$.  
% Under the usual assumption, the data in each class form a differentiable manifold. For a given $k$ and any ${\bf X} \in {\cal M}^{(k)}$, let ${\cal C}({\bf X})$ be the convex hull of the $k$ points in ${\bf X}$.
% %and $y_1, y_2, \ldots, y_k$ be the distinct labels of the $k$ points. 
% Then the set of all policies $\lambda$ such that 
% ${\bf X}\lambda \in {\cal C}({\bf X})\setminus {\cal M}$ is 
% $\Lambda^*({\bf X})$. Thus $\Lambda^*({\bf X})$ is smooth almost everywhere. Then it can be approximated by a network $\pi_k$ (having finite but possibly large capacity) arbitrarily well. For the same reason, there exists a finite-capacity intrusion discriminator that can classify arbitrarily well the points from ${\cal M}$ and the poins ${\bf X}\lambda$ (with 
% $\lambda$ drawn from $\pi_k({\bf X})$). This will then give arbitrarily low cross entropy loss.  \hfill $\Box$

%Due to this result, one may use the achievable value of ${\cal L}_{\rm intr}$ to measure how well the data supports MixUp without causing intrusion. 

\section{Implementation of AdaMixUp}

\subsection{Implementation of $\{\pi_k\}$}

Instead of constructing $k_{\rm max}-1$ networks $\pi_k$'s, we in fact implement these networks recursively using a single network $\pi_2$. For this purpose, we recursively parametrize $\Lambda({\bf X}) \in {\mathbb S}_k$, which we will rewrite as $\Lambda_k({\bf X})$ for clarity. 

For an ${\bf X}\in {\cal X}^{(k)}$, we may express it as $[{\bf X}^{(k-1)}, x_k]$, where 
${\bf X}^{(k-1)}$ is the sub-matrix of ${\bf X}$ containing the first $k-1$ columns.  Then， for $k>2$, we parametrize 
\begin{equation}
\begin{split}
\Lambda_k({\bf X}): = 
&
\left\{
\left[
\begin{array}{c}
\gamma \lambda\\
1-\gamma
\end{array}
\right]:
\right.
\lambda \in \Lambda_{k-1}\left({\bf X}^{(k-1)}\right), 
\\
&\left.
\left[
\begin{array}{c}
\gamma\\
1-\gamma
\end{array}
\right]
\in \Lambda_2\left(
[\lambda {\bf X}^{(k-1)}, x_k]
\right)
\right\}
\end{split}
\end{equation}

This then allows the use of single network $\pi_2$ to express $\Lambda_k({\bf X})$ for all ${\bf X}$ and all $k$.

Specifically, a Fold-2 mixing policy is parameterized by a single number $\gamma$. For each input ${\bf X}\in {\cal X}^{(2)}$, $\pi_2$ returns two positive values $\alpha$ and $\Delta$ with $\alpha+\Delta \le 1$, which specifies $\Lambda({\bf X})$ as the interval $(\alpha, \alpha+\Delta)$ as the range for $\gamma$.  To this end, the last layer of the $\pi_2$ 
network is implemented as a softmax function, which generates a triplet of values $(\alpha, \Delta, \alpha')$ with sum of one, where the third element $\alpha'$ is discarded. 
Using network $\pi_2$, we generate a set of Fold-2 mixed examples by repeatedly drawing a pair of samples in ${\cal D}$ and mixing them using a random policy drawn from policy region generated by $\pi_2$ on the pair of samples. 

To generate mixed samples with an increased maximum fold number, we consider the original examples and previously generated mixed examples as the new training set ${\cal D}_{\rm new}$. We then repeat the following process: draw a pair of samples, one from ${\cal D}_{\rm new}$ and the other from ${\cal D}$; submit the pair to network $\pi_2$ to generate a policy region; draw a random policy from the region and obtain a new sample by mixing the pair of samples using the drawn policy.

\subsection{Reparametrization Trick}
To allow the gradient signal to back-propagate through the policy sampler, a reparametrization trick similar to that of ~\cite{DBLP:journals/corr/KingmaW13} is used. Specifically, 
drawing $\gamma$ from $(\alpha, \alpha+\Delta)$ is implemented by drawing 
$\epsilon$ from the uniform distribution over $(0, 1)$, then let $\gamma:=\Delta \cdot \epsilon + \alpha$.

\begin{figure}
\begin{center}
%\begin{tabular}{c}
\scalebox{0.7}{
\begin{tikzpicture}[-latex ,auto ,node distance =2 cm and 2 cm, on grid ,
semithick ,
state/.style ={ circle ,top color =white , bottom color = red!20 ,
draw, red , text=red , minimum width =0.05cm},
box/.style ={rectangle ,top color =white , bottom color = processblue!20 ,
draw, processblue , text=blue , minimum width =0.1cm , minimum height = 0.1cm}]

 \node[draw,rectangle] (a) {$x_1$};
 \node[draw,rectangle,right=1cm of a] (c) {$x_2$};

 % \node[draw,rectangle,below right = 1cm and 1cm of a,text width=1.5cm] (p) {Policy   Region   Generator}; 
  \node[draw,rectangle,above right= 2cm and 0.5cm of a,text width=1.5cm] (p) {Policy   Region   Generator}; 
 \node[inner sep=0,minimum size=0,right = 1.4cm of p] (k) {$\gamma$};% invisible node
 \node[draw,rectangle,right = 2.3cm of k] (b) {$\hat{x}=\gamma x_1 +(1-\gamma) x_2$};
 
 \node[draw,rectangle,above right=1cm and 1cm of b,minimum height=0.6cm] (c11) {$x_1$};
 \node[draw,rectangle,right=0.6cm of c11,minimum height=0.6cm] (c12) {$x_2$};
 \node[draw,rectangle,right=0.56cm of c12,minimum height=0.6cm] (c13) {$\hat{x}$};

\node[draw,rectangle,below right=1cm and 1cm of b] (c21) {$\hat{x} / +$};
 \node[draw,rectangle,right=1.02cm of c21] (c22) {$x_1 / -$};
 \node[draw,rectangle,right=1.05cm of c22] (c23) {$x_2 / -$};
 \node[draw,rectangle,right=1.05cm of c23] (c24) {$x_3 / -$};
   \node[draw,rectangle,right = 2cm of c24,text width=1.6cm] (id) {Intrusion Discriminator};
 \node[draw,rectangle,below = 1cm of c24] (d) {$x_3$};
  \node[draw,rectangle,above of = id,text width=1.6cm] (cl) {Classifier};

\path (a) edge (p);
\path (c) edge (p);
\path (p) edge (k);
\path (k) edge (b);
\path (b) edge (c11);
\path (b) edge (c21);
\path (c13) edge (cl);
\path (c24) edge (id);
\path (d) edge (c24);

\end{tikzpicture}
}
%\end{tabular}
\end{center}
\vspace{-0.3cm}
\caption{\label{fig:schema} Fold-2 AdaMixUp for a single triplet $(x_{1}, x_{2}, x_{3})$.} Each batch is implemented to contain multiple such triplets. ``+'' and ``-'' indicate positive and negative examples, respectively. 
\label{schema}
\end{figure}
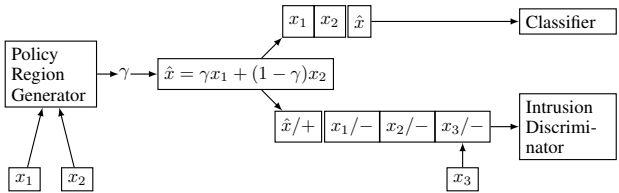

\subsection{Implementation of $\varphi(\cdot)$}
The network $\varphi$ is
a neural network that classifies original examples from the mixed examples. In our implementation, $\varphi$ shares with the classifier network all but the final layers. The final layer of $\varphi$ is a simple logistic regression binary classifier.

\subsection{Training}
The  network $\varphi$ and $\pi_2$ are trained jointly, where we iterate over minimizing ${\cal L}_{\cal D}+ {\cal L}_{{\cal D}'}$ and minimizing ${\cal L}_{\rm intr}$ in (\ref{eq:lossAdaMixUp}) via mini-batched SGD. 
A Fold-2 AdaMixUp is illustrated in Figure~\ref{schema}.

\section{Experiments}
\subsection{Data Sets and Experimental Settings}
We evaluate  AdaMixUp  on eight   benchmarks. %\footnote{The code of our implementation of AdaMixUp will be made available at \url{https://github.com/SITE5039/AdaMixUp/}}. 
%\begin{itemize}
%\item 
\noindent {\em MNIST} is the popular digit (1-10)  recognition dataset with 60,000 training and 10,000 test gray-level, 784-dimensional images.
%\item 
%\noindent 
{\em Fashion} is an image recognition dataset having the same scale as
MNIST, containing 10 classes of fashion product pictures.
%\item 
%\noindent 
{\em SVHN} is the Google street view house numbers recognition data set with 73,257 digits 
(1-10) 32x32 color images for training, 26,032 for testing, and 531,131 additional, easier samples. We did not use the additional images.
%\item 
%\noindent 
{\em Cifar10} is an image classification dataset with  10 classes, %(airplanes, cars, birds, cats, deer, dogs, frogs, horses, ships, and trucks)
 50,000 training  and 10,000 test samples. 
%\item 
%\noindent 
{\em Cifar100} is similar to CIFAR10 but with 100 classes and 600 images each.
%\item 
%\noindent 
{\em Cifar10-S} and {\em Cifar100-S} are respectively Cifar10 and Cifar100 reduced to containing only 20\% of the training samples. 
%\item 
{\em ImageNet-R} is the ImageNet-2012 classification dataset~\cite{RussakovskyDSKSMHKKBBF14} with  1.3 million training images, 50,000 validation images, and 1,000 classes. We follow the data processing approaches used in Mixup~\cite{mixup17}, except that the crop size is 100x100 instead of 224x224 due to our limited computational resources.  We report both top-1 and top-5 error rates.

In our experiments, the Intrusion Discriminator,   Classifier, and Policy Region Generator  have the same network architecture, and the first two share the same set of parameters. 
We test AdaMixUp on two types of baseline networks: a three layers CNN as  implemented in~\cite{wu2016tensorpack} as the baseline  for easier tasks MNIST and Fashion, and a ResNet-18 as implemented  in~\cite{Zagoruykocode} for the other six more  difficult tasks. All models examined are trained using mini-batched backprop, as specified in ~\cite{wu2016tensorpack} and ~\cite{Zagoruykocode},     for 400 epochs.   Each reported performance value (accuracy or error rate) is the median of the performance values obtained in the final 10 epochs.  

\subsection{Predictive Performance}
\label{acc}
Our first experiment compares  Fold-2 AdaMixUp (i.e., mixing image pairs) to  two networks: the baseline networks (i.e., 3-layer CNN for MNIST and Fashion and ResNet-18 for the rest  datasets) and  MixUp on the baseline networks. We present the error rates obtained by the Baseline, MixUp, and AdaMixUp, along with the relative error reduction of AdaMixUp over the Baseline,  in Table~\ref{tab:accuracy:resnet}.  

Table~\ref{tab:accuracy:resnet} indicates that the AdaMixUp  outperforms  both the Baseline and MixUp on all the eight testing datasets. Also, the relative error reduction  of AdaMixUp over the Baseline is at least 5.7\%, and with a large margin  in some cases. For example,  for the SVHN and Cifar10 datasets, the relative error reduction achieved by AdaMixUp  are over 30\%. 

Table~\ref{tab:accuracy:resnet} also suggests that, with the default $\alpha$ value, MixUp failed to improve the baseline systems'  accuracy on four  out of the eight datasets, namely the MNIST, Cifar10-S,  Cifar100-S, and Imagenet-R datasets, as underlined in Table~\ref{tab:accuracy:resnet}, suggesting over-regularization or under-fitting. %In contrast, the AdaMixUp was able to achieve a relative accuracy gain of 5.77\%, 10.81\%, and 6.15\% for these three datasets, respectively.
\begin{table}[h]
  \centering
    \resizebox{\columnwidth}{!}{
\begin{tabular}{l|r|r|r|c}\hline
Data Set& Baseline&MixUp &Ada&Relative \\ 
& & &MixUp&  Impro. (\%)\\ \hline
mnist&0.52&\underline{0.57}&0.49&5.77\\
fashion&7.37&6.92&6.21&15.74\\
svhn&4.50&3.80&3.12&30.67\\
cifar10&5.53&4.24&3.52&36.35\\
cifar100&25.6&21.14&20.97&18.09\\
cifar10-S&7.68&\underline{7.88}&6.85&10.81\\
cifar100-S&28.47&\underline{29.39}&26.72&6.15\\
ImageNet-R top1&53.00&\underline{54.89}&49.17&7.22\\
ImageNet-R top5&29.41&\underline{31.02}&25.78&12.34\\

%ImageNet-R&&&&\\
 \hline
\end{tabular}}
  \caption{Error rates (\%) obtained by the testing methods.}
  \label{tab:accuracy:resnet} 
\end{table}

In Table~\ref{tab:accuracy:para}, we also present the $\alpha$ and $\Delta$ values learned by the Policy Region Generator, along with the  losses of  both the Intrusion Discriminator and  the Classifier of the AdaMixUp method. Results in Table~\ref{tab:accuracy:para}  indicate that the values of  $\alpha$ and  $\Delta$ vary for different datasets; the former ranges from 0.4 to 0.9 but the range for $\Delta$ is much smaller, with values between  0.010 and 0.035.  
Noting that the intrusion losses 
are fairly close to 0 on all the datasets, suggesting  %according to Lemma \ref{lem4} 
that these datasets well support Fold-2 AdaMixUp under the proposed structure of Policy Region Generator.

\begin{table}[h]
  \centering
\begin{tabular}{l|c|c|c|c}\hline
Ada values&$\alpha$&$ \Delta$&${\cal L}_{\rm intr}$&${\cal L}_{\cal D} + {\cal L}_{{\cal D}'}$ \\ \hline
mnist&0.497&0.010&0.002&0.201\\
fashion&0.511&0.011&0.003&0.290\\
svhn&0.924&0.011&0.002&0.116\\
cifar10&0.484&0.029&0.003&0.371\\
cifar100&0.484&0.034&0.002&0.679\\
cifar10-S&0.486&0.027&0.003&0.479\\
cifar100-S&0.482&0.035&0.007&0.712\\
ImageNet-R&0.493&0.004&0.038&2.382\\
%ImageNet-R&&&&\\
 \hline
\end{tabular}
  \caption{Fold-2 AdaMixUp: Mean $\alpha$, $\Delta$, and  training losses.}
  \label{tab:accuracy:para} 
\end{table}

\subsection{Training Characteristics}
\begin{figure}[h]
\caption{The mean of $\alpha$ (left) and the mean of $\Delta$ (right) in Fold-2 AdaMixUp on Cifar10.}
	\centering
		\includegraphics[width=3.5in,height=1in]{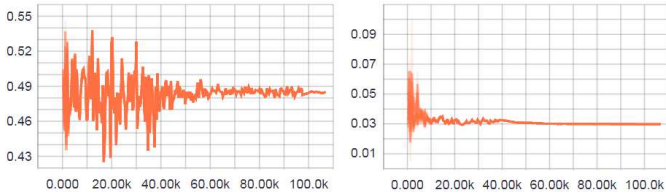}%schema.eps}
	\label{fig:alpha}
\end{figure}

Figure~\ref{fig:alpha} depicts the behavior of Policy Region Generator in Fold-2 AdaMixUp  over training iterations. It appears that the Policy Region Generator initially explores a wide range of policy space before settling down at around 40K iterations (when the means of $\alpha$ and $\Delta$ both stabilize, at 0.48 and 0.03 respectively). 

\begin{figure}[h]
\caption{Fold-2 AdaMixUp on Cifar10: Mean of mixing policy $\gamma$  (left) and training/testing error rates (right).} 
	\centering
\includegraphics[width=3.42in]{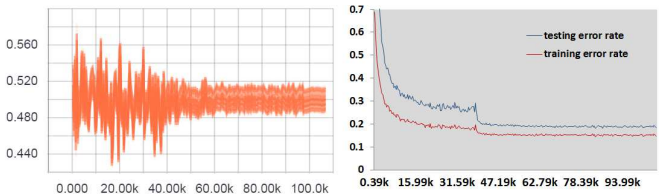}%schema.eps}
	\label{fig:weights}
\end{figure}

Figure~\ref{fig:weights} shows the 
average of the drawn mixing policy $\gamma$ in Fold-2 AdaMixUp during training iterations and the corresponding classification performance of the model. The range of $\gamma$ appears to stabilize at around 40K iterations (left plot), consistent with the observation in Figure \ref{fig:alpha}. Interestingly, also at the same time, the classification error drops and begins to stabilize 
(right plot).

\begin{figure}[h]
\caption{Mixed images (top row) in Fold-2 AdaMixUp from the original images (bottom 2 rows) in MNIST.}
	\centering
		\includegraphics[width=3.3062in]{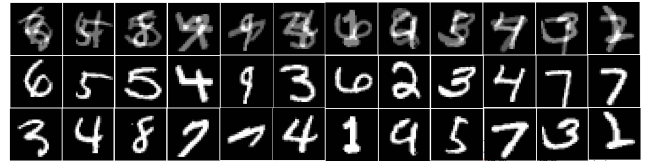}%schema.eps}
	\label{fig:localization2}
\end{figure}

Figures~\ref{fig:localization2} shows some typical mixed images in  Fold-2 AdaMixUp  and their corresponding original images in MNIST. We note that these mixed images obviously distinguish themselves from the original images. Thus they do not intrude into the data manifold.

\subsection{Impact of Mixing Fold}
%This section evaluates the impact the fold number in the AdaMixUp method. 
Table~\ref{tab:accuracy:nfold} presents the error rates obtained by Fold-3 and Fold-4 AdaMixUp on the seven testing datasets (excluding ImageNet due to limited computational resources). We see that increasing the mixing fold from 2 to higher values, the performance of AdaMixUp improves. This is due to stronger constraints are imposed to the model and hence results in stronger regularization. One however expects that imposing stronger constraints makes the model more difficult to fit and further increasing mixing fold may eventually lead to diminishing gain and even under-fitting.  This is hinted by Table \ref{tab:accuracy:nfoldcost}, where the averages of learned  mixing parameters and the training losses of the Fold-3 AdaMixUp are presented.

In Table ~\ref{tab:accuracy:nfoldcost}, we see that the intrusion losses of Fold-3 AdaMixUp are much higher than those of Fold-2 AdaMixUp (Table \ref{tab:accuracy:para}). This implies that in higher-fold AdaMixUp, it is more difficult for the Policy Region Generator to carve a good policy region that is not regarded by the Intrusion Discriminator as causing intrusion. This difficulty is also suggested by the high $\alpha_2$ values in Table \ref{tab:accuracy:nfoldcost} (often close to 1). They indicate that in these cases, mixing only occurs ``slightly''.

\begin{table}[h]
  \centering
  
\begin{tabular}{l|c|c|c}\hline
AdaMixUp& Fold-2&Fold-3&Fold-4  \\\hline
%& &&  \\ 
mnist&0.49&0.42&0.41 \\
fashion&6.21&5.88&5.71\\
svhn&3.12&2.98&2.91\\
cifar10&3.52&3.25&3.21\\
cifar100&20.97&20.68&20.29\\
cifar10-S&6.85&6.14&5.96\\
cifar100-S&26.72&25.72&25.49\\
%ImageNet&&&\\
 \hline
\end{tabular}
  \caption{Error rates (\%) of Fold-3 and Fold-4 AdaMixUp.}
  \label{tab:accuracy:nfold} 
\end{table}

\begin{table}[h]
  \centering
  \resizebox{\columnwidth}{!}{
\begin{tabular}{l|c|c|c|c}\hline
Fold-3& $\alpha1$/$\Delta1$&$\alpha2$/$\Delta2$&${\cal L}_{\rm intr}$&${\cal L}_{\cal D} + {\cal L}_{{\cal D}'}$\\\hline
%AdaMixUp&&&Cost&Cost\\
mnist&0.533/0.013&0.662/0.013&0.002&0.337\\
fashion&0.517/0.015&0.678/0.019&0.024&0.327\\
svhn&0.637/0.011&0.943/0.020&0.028&0.482\\
cifar10&0.516/0.028&0.938/0.041&0.003&0.403\\
cifar100&0.121/0.010&0.959/0.015&0.004&0.305\\
cifar10-S&0.504/0.039&0.849/0.068&0.028&0.505\\
cifar100-S&0.639/0.011&0.943/0.018&0.023&0.542\\
%ImageNet&&&&\\
 \hline
\end{tabular}}
  \caption{Fold-3 AdaMixUp: average training losses and policy region parameters. $(\alpha_1, \Delta_1)$: initial Fold-2 mixing parameter; $(\alpha_2, \Delta_2)$: further mixing parameter.}
  \label{tab:accuracy:nfoldcost} 
\end{table}

\subsection{Sensitivity to Model Capacity} 
\begin{figure}[h]
\caption{Error rates  obtained by ResNet-18 and AdaMixUp on Cifar100, when varying the number of filters  (green curves) and varying the size of the  samples (red curves). }
\label{filtersize}
	\centering
    		\includegraphics[height=1.69in]{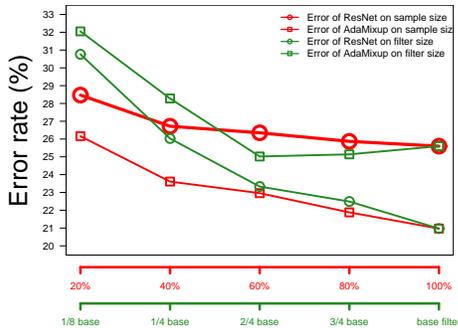}%schema.eps}	
\end{figure}

We vary the number of filters  on each layer of the ResNet-18  with half quarter, quarter, half, and three-quarter of the original number of filters (denoted as base filter), and present the test error rates  on  Cifar100 in Figure~\ref{filtersize} (green curves). 

The green curves in Figure~\ref{filtersize} indicate that the Fold-2 AdaMixUp  benefits from large number of filters deployed: the accuracy keeps improving while  enlarging the number of filters. %Intuitively, stronger model enables the Policy Region Generator to explore more  policy regions not colliding with the  data manifold, thus further benefiting the AdaMixUp.  
On the contrary, the ResNet-18 received dramatically accuracy improvement before the number of filters is half of the base but obtained no further improvement after.  %Similar patterns can be seemed in the loss curves. 

\subsection{Effect of Data Size}
We down sample the  Cifar100 data with 20\%, 40\%, 60\% and 80\% of the  training samples, and present the test error rates obtained by Fold-2 AdaMixUp in  Figure~\ref{filtersize}  (red curves). 

The red curves in Figure~\ref{filtersize} indicate that more training samples help with both the ResNet-18 and AdaMixUp, but the accuracy gap between the two methods are widening while increasing the training sample size. 
%We hypothesis that more data samples give rise to more feasible (no manifold intrusion) mixing policies for the  generation of synthetic samples, thus resulting in additional model constraints for the regularization.

\subsection{Benefit of the Intrusion Discriminator}
%We further evaluate the benefit of deploying the Intrusion Discriminator. 
Table~\ref{tab:accuracy:resnet:nointrusion}  lists the test error rates  obtained by the Fold-2 AdaMixUp with ResNet-18 on  the Cifar10 and Cifar100 data, but without  the Intrusion Discriminator. 
\begin{table}[h]
  \centering
  %\resizebox{\columnwidth}{!}{
%\begin{tabular}{l|c|c|c|c}\hline
%Data Set& Baseline&MixUp&Ada&AdaMixUp w/o\\ 
% &ResNet-18&&MixUp&Intru. Discr.\\ \hline
%cifar10&5.53&4.24&3.52&3.83\\
%cifar100&25.6&21.14&20.97&24.75\\
\begin{tabular}{l|c|c|c}\hline
Data Set& Baseline&Ada&AdaMixUp w/o\\ 
 &ResNet-18&MixUp&Intru. Discr.\\ \hline
cifar10&5.53&3.52&3.83\\
cifar100&25.6&20.97&24.75\\

 \hline
\end{tabular}
%}
  \caption{Error rates obtained by  Fold-2 AdaMixUp without the Intrusion Discriminator on the Cifar10 and Cifar100.}
  \label{tab:accuracy:resnet:nointrusion} 
\end{table}
Table~\ref{tab:accuracy:resnet:nointrusion} shows that excluding the Intrusion Discriminator component, the AdaMixUp method  was still able to improve the accuracy over the baseline, %(from 5.53\% to 3.83\% and from 25.6\% to 24.75\%, respectively, for Cifar10 and Cifar100), 
but obtained much lower accuracy than that of including the Intrusion Discriminator unit. % (with 3.52\% and 20.97\% respectively). 
%\textcolor{red}{
%Without the Intrusion Discriminator, manifold intrusion can easily occur and thereby resulting in degrading the predictive performance of the model.  Interestingly, when comparing to MixUp, the AdaMixUp without the Intrusion Discriminator seemed to have more chance to collide into the manifold on the Cifar10 than that on the Cifar100 (may due to the large label space in Cifar100), resulting in worse accuracy than MixUp on Cifar100, but better accuracy on Cifar10. In other words, in the Cifar100 case, the uniformly randomly chosen mixing policies in  MixUp are better than the narrowed down policy regions learned by AdaMixUp. }

\subsection{Interpolating on Hidden Layer}
We also evaluate  an alternative structure for  AdaMixUp, where mixing happens  on the layer before the final softmax layer in ResNet-18. Our  results suggest that the test errors of the Fold-2 AdaMixUp increased  dramatically due to the high cost of the  Intrusion Discriminator. The error rates increased from 20.97\% to 22.21\% and from 3.52\% to 4.94\%, respectively for Cifar100 and Cifar10. Notably, both have large intrusion loss of around 0.49, which may due to the fact that perceptual similarity for images in the hidden embedding space  may not as distinguishable as that in the input space. As a result, the Policy  Generator failed to find good mixing policies to avoid colliding into the  data manifold.

\section{Related Work}
%Data augmentation lies at the heart of many successful deep learning applications due to its capability of leading to improved model generalization, but 
%Common data augmentation methods have been designed based on substantial domain knowledge~\cite{726791,SimonyanZ14a,abs170804896}, relied on specific network architectures~\cite{Gastaldi17}, or  leveraged feedback signals to search the optimal augmentation strategies~\cite{DBLP:journals/corr/abs180509501}. Our method excludes those requirements, and only leverages a simple linear interpolation for data augmentation. 

%Our work  closely relates to approaches linearly interpolating   examples and labels~\cite{mixup17,ManifoldMixUp,DBLP:journals/corr/abs171110284}. Nevertheless, these approaches   depends on the correct user-predefined mixing policies.  Also, their interpolation typically lies along the lines of sample pairs. On the contrary,  our approach automatically learns the  mixing policy regions and  benefits from mixing  multiple images. 

Common data augmentation methods have been designed based on substantial domain knowledge~\cite{726791,SimonyanZ14a,AmodeiABCCCCCCD15,abs170804896}, relied on specific network architectures~\cite{Gastaldi17,DBLP:journals/corr/abs170804552,ttbabs18020237}, or  leveraged feedback signals to search the optimal augmentation strategies~\cite{DBLP:journals/corr/LemleyBC17,DBLP:journals/corr/abs180509501}. 
Our method excludes those requirements, and only leverages a simple linear interpolation for data augmentation. 

Our work  closely relates to approaches linearly interpolating   examples and labels~\cite{mixup17,articleDeVries,ManifoldMixUp,DBLP:journals/corr/abs171110284}. Nevertheless, these approaches   depends on the correct user-predefined mixing policies.  Also, their interpolation typically lies along the lines of sample pairs. 
On the contrary,  our approach automatically learns the  mixing policy regions and  benefits from mixing  multiple images.

%\textcolor{red}{Should we say something about Manifold regularization?}

\section{Conclusions and Outlook}

This work justifies the effectiveness of MixUp from the perspective  of out-of-manifold regularization. We also identify the inherent problem with standard MixUp in causing manifold intrusion. This motivates us to develop AdaMixUp, which generalizes MixUp to higher-fold mixing policies and automatically learns the policy regions that avoid manifold intrusion.

To us, this work is only the beginning of a rich research theme. The justification of MixUp and AdaMixUp we provide thus far is primarily qualitative. It is desirable to quantitatively characterize the generalization capability of AdaMixUp. Moreover, the out-of-manifold regularization perspective potentially opens a door to new regularization techniques.  Beyond local linearization, we believe that there is a rich space of other possibilities.

%\section*{Acknowledgments}

%The authors wish to thank the anonymous reviewers for their valuable comments, some of which are likely to help furthering this work. 

%This work is supported partly by China 973 program (2015CB358700), by the National Natural Science Foundation of China (61772059), and by the Beijing Advanced Innovation Center for Big Data and Brain Computing.

\bibliographystyle{aaai}
\bibliography{reference}
\end{document}